\pdfoutput=1

\documentclass[11pt]{article}
\usepackage[preprint]{acl}

\usepackage{url} 

\usepackage{times}
\usepackage{latexsym}
\usepackage{amsmath}
\usepackage{graphicx}
\usepackage{float}
\usepackage{caption}
\usepackage{subcaption}
\usepackage{amssymb}

\usepackage[T1]{fontenc}
\usepackage[utf8]{inputenc}

\usepackage{microtype}

\usepackage{inconsolata}
\usepackage{booktabs}
\usepackage{xcolor}

\newcommand{\newshortname}{CounterCurate}

\title{\newshortname{}: Enhancing Physical and Semantic Visio-Linguistic Compositional Reasoning via Counterfactual Examples}

\author{
Jianrui Zhang\thanks{Equal Contribution.}$^{1}$ \quad Mu Cai$^{*1}$  
  \quad  Tengyang Xie$^{1,2}$  \quad   Yong Jae Lee$^{1}$\\
  \vspace{-0.5em}\\
  \texttt{\{harrisz,mucai,tx,yongjaelee\}@cs.wisc.edu}\\
  \vspace{-0.8em}\\
$^{1}$University of Wisconsin--Madison~~~~~~$^2$Microsoft Research 
\\
  \vspace{-0.8em}\\
  \href{https://countercurate.github.io}{https://countercurate.github.io}
}

\begin{document}
\maketitle

\begin{abstract}

We propose \newshortname{}, a framework to comprehensively improve the visio-linguistic compositional reasoning capability for both contrastive and generative multimodal models. In particular, we identify two critical under-explored problems: the neglect of physically grounded reasoning (counting and position understanding) and the potential of using highly capable text and image generation models for semantic counterfactual fine-tuning. Our work pioneers an approach in addressing these gaps.

We first spotlight the near-chance performance of multimodal models like CLIP and LLaVA in physically grounded compositional reasoning. We then apply simple data augmentation using the grounded image generation model GLIGEN to generate fine-tuning data, resulting in significant performance improvements: +33\% and +37\% for CLIP and LLaVA, respectively, on our newly curated Flickr30k-Positions benchmark.
Moreover, we exploit the capabilities of high-performing text generation and image generation models, specifically GPT-4V and DALLE-3, to curate challenging semantic counterfactuals, thereby further enhancing compositional reasoning capabilities on benchmarks such as SugarCrepe, where \newshortname{} \textbf{outperforms GPT-4V}. 

To facilitate future research, we release our code, dataset, benchmark, and checkpoints at \url{https://countercurate.github.io/}.

\end{abstract}

\section{Introduction}

\begin{figure}[t]
\centering
\includegraphics[width=\linewidth]{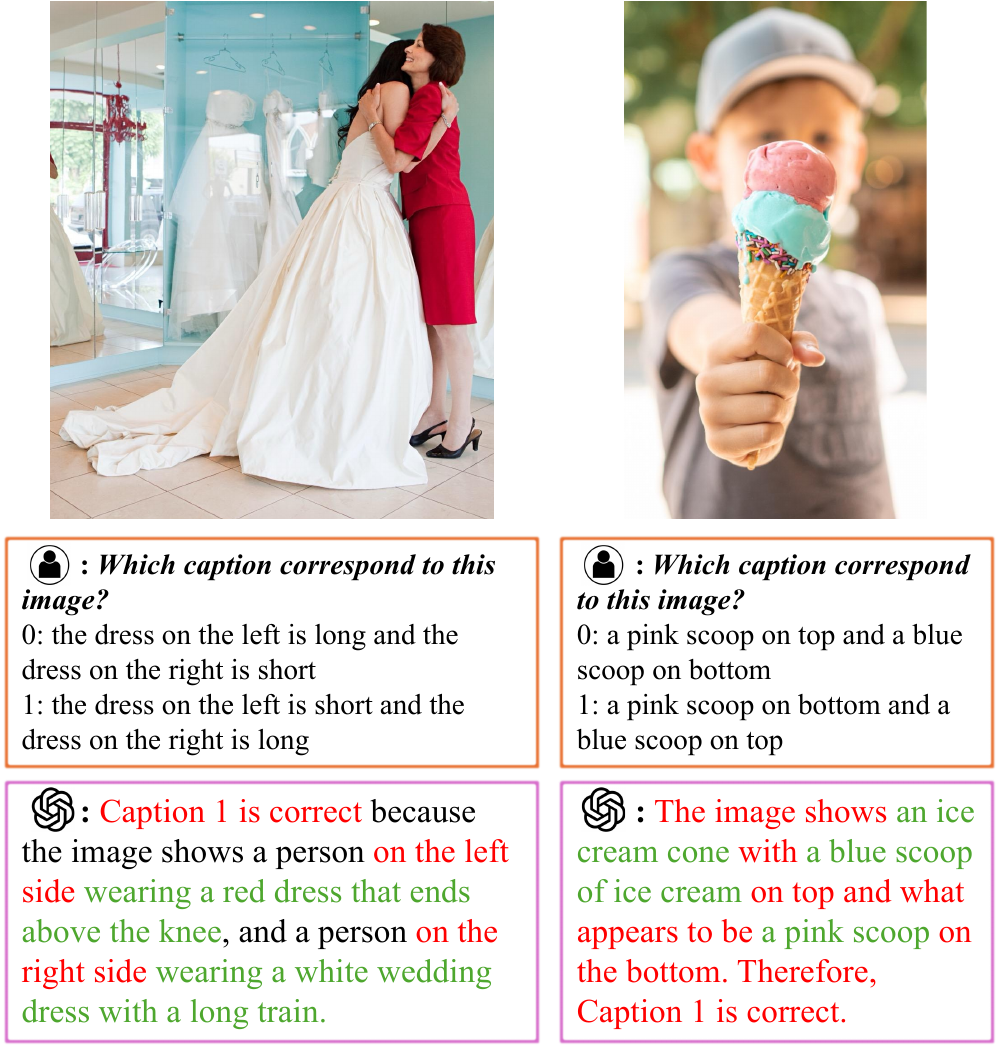}
\caption{Representative examples of GPT-4V failure cases. In both questions, GPT-4V correctly identifies all objects in question, but chooses the wrong answer because it fails to distinguish between either left and right (the left question) or up and down (the right question).}

\label{fig:gpt4v-fail}
\end{figure}

Large language models such as ChatGPT~\cite{chatgpt}, GPT4~\cite{openai2023gpt4}, and LLaMA~\cite{touvron2023LLaMA} have demonstrated remarkable knowledge and reasoning abilities. Recently, large multimodal models (LMMs) \cite{radford2021learning, openai2023GPT4V} further leverage large-scale image-text pairs for improving visio-linguistic understanding capabilities. However, these models exhibit subpar performance in compositional reasoning \cite{diwan2022winoground, hsieh2023sugarcrepe}, such as differentiating between semantic distractors like ``\textit{white shirts and black pants}'' and ``\textit{black shirts and white pants}''. Current research either concentrates on curating evaluation benchmarks~\cite{diwan2022winoground, hsieh2023sugarcrepe, le2023cococounterfactuals} or enhancing compositional reasoning abilities by creating rule-based counterfactual fine-tuning data~\cite{yuksekgonul2023when, le2023cococounterfactuals}.

Currently, the majority of research focuses on semantic compositional reasoning, leaving another fundamental problem, physically grounded compositional reasoning, under-explored. This involves tasks such as counting and distinguishing left/right and up/down positions between objects. For example, in Figure~\ref{fig:gpt4v-fail}, GPT-4V \cite{openai2023GPT4V} made the wrong choice even though it identified all objects in question, demonstrating capable semantical yet weak physical compositional reasoning.\footnote{We were able to reproduce our findings of GPT's weakness in multiple fashions: the failure cases themselves were initially discovered via manual testing before the APIs were available; we were able to both use the 2023-07-01 Azure API and the website version of GPT4-V to recreate many of the failure cases as well.}

\begin{figure*}[t]
\centering
\includegraphics[width=\linewidth]{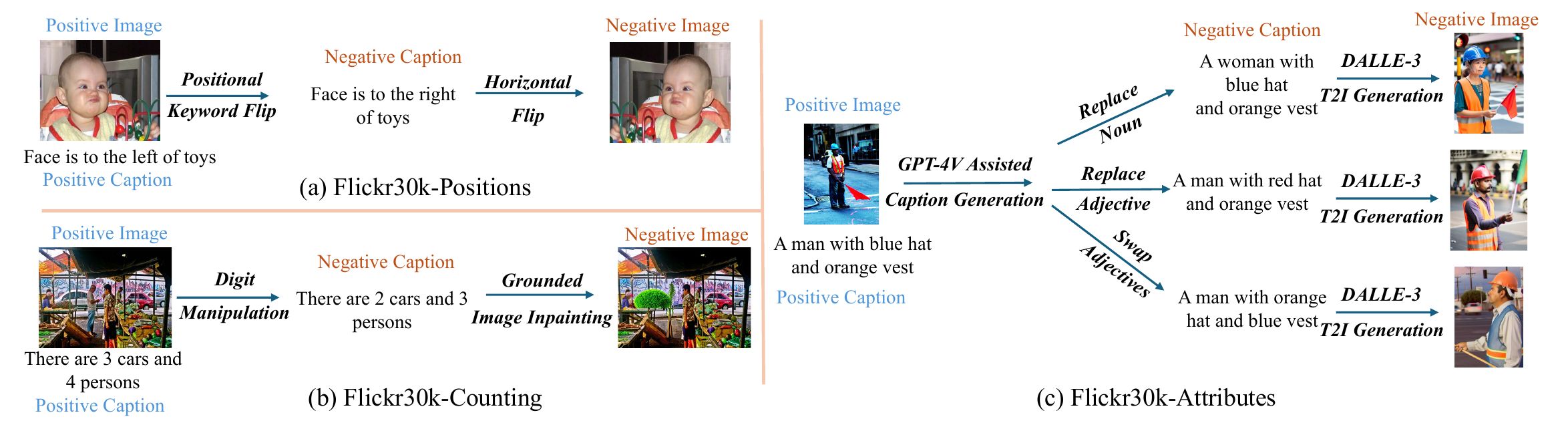}

\caption{The data curation pipeline of \newshortname{}. Given a positive image-caption pair, we first generate the negative captions, based on which we curate the negative images using the most suitable approach. Specifically, (a) for Flickr30k-Positions (left/right), we flip the positional keyword before conducting the horizontal flip for the image; (b) for Flickr30k-Counting, we manipulate the digit before applying grounded image inpainting~\cite{li2023gligen} as the negative image; (c) for Flickr30k-Attributes, we first leverage GPT-4V~\cite{openai2023GPT4V} to generate reasonable hard negative captions for replacing the noun, replacing the adjective, and swapping the adjectives. Then we leverage  DALLE-3~\cite{openai2023DALLE3} to generate coherent images. }

\label{fig:data_curation}
\end{figure*}

With such an observation, it is not surprising to find that both contrastive models like CLIP~\cite{radford2021learning} and generative models like LLaVA~\cite{liu2023llava} deliver near random chance performance on our newly curated benchmarks. We hypothesize that modern LMMs are largely oblivious to positional differences. To address this, we generate counterfactual image-text pairs using simple methods, such as horizontal flipping for left/right distinctions, as well as by leveraging a grounded image generation model, GLIGEN~\cite{li2023gligen}, to accurately replace/remove objects of interest for up/down and counting tasks. Our approach shows significant improvements such as 33+\% for CLIP and 37+\% for LLaVA on our Flickr30k-Positions benchmark.

Existing methods also do not fully utilize the capabilities of powerful generative models when creating semantic counterfactual fine-tuning data. We argue that the key to enhancing compositional reasoning capabilities lies in the careful curation of accurate and sufficiently challenging image and text counterfactuals. For example, augmented negative captions with linguistic rules used by recent fine-tuning approaches \citep[e.g.,][]{yuksekgonul2023when, le2023cococounterfactuals} can unintentionally follow unnatural language distributions, which are easily distinguishable by a text-only model without any image evidence \citep{lin2023visualgptscore}.

To overcome this, we utilize high-performance text generation model GPT-4V~\cite{openai2023GPT4V} and image generation model DALLE-3~\cite{openai2023DALLE3} to curate reasonably and sufficiently difficult negatives. Our method empirically demonstrates a significant performance boost by fine-tuning CLIP and LLaVA using our data generation pipeline on benchmarks such as SugarCrepe, where we surpass NegCLIP and GPT-4V.

Our contributions are summarized as follows:
\begin{itemize}
\item We systematically study physically grounded compositional reasoning, including positional understanding and counting, and highlight the near random performance of multimodal models, including CLIP and LLaVA, on our newly curated benchmarks.
\item We significantly improve physical reasoning capabilities by utilizing simple data augmentation techniques and a grounded image generation model, GLIGEN, to generate counterfactual images and captions.
\item We employ the most capable image and text generation models to generate semantically counterfactual image/text pairs, further enhancing performance over SOTA methods.
\end{itemize}

\section{Related Work}

\subsection{Large Multimodal Models}

Multimodal models aim at connecting multiple data modalities. For visio-linguistic tasks, multimodal models are constructed either as contrastive~\cite{radford2021learning, jia2021scaling} or generative~\cite{openai2023gpt4, alayrac2022flamingo, liu2023llava}. Recent large multimodal models are trained with large-scale pretraining data and billions of trainable parameters, allowing them to excel at generalization. Contrastive models such as CLIP~\cite{radford2021learning} connect images and texts by aligning two modality-specific encoders using contrastive loss. Generative models such as LLaVA~\cite{liu2023llava,liu2023improved} embed images into a decoder-only language model as prefix tokens and utilize next-token prediction loss to align the two modalities. Though such models demonstrate impressive capabilities across various vision-language tasks such as retrieval and visual question answering, they perform less desirably over compositional reasoning~\cite{diwan2022winoground, hsieh2023sugarcrepe}. 

\subsection{Compositional Reasoning}

Compositional reasoning~\cite{diwan2022winoground} in vision and language tasks involves evaluating models' ability to understand and manipulate complex ideas by breaking them down into simpler, constituent components before recombining them in novel ways. A typical example can be distinguishing ``blank pants and white shorts'' versus ``white pants and blank shorts''.  
Winoground~\cite{diwan2022winoground} is the pioneering benchmark for compositional reasoning, composed of 400 items, each with two images and two corresponding captions. Given an image, multimodal models are asked to find the matching caption from the provided two options, and vice versa. SugarCrepe~\cite{hsieh2023sugarcrepe} and SeeTrue~\cite{yarom2023what} further scale the distractive captions by leveraging language models for automated generation.

NegCLIP~\cite{yuksekgonul2023when} attempts to improve compositional reasoning via fine-tuning CLIP with perturbed captions as the negative text samples in contrastive learning. However, such permuted captions can be easily detected by a text-only transformer~\cite{lin2023visualgptscore}, demonstrating that a more effective approach is needed to further enhance compositional reasoning. DAC~\cite{doveh2024dense} leverages dense and aligned captions for better positive captions. SGVL~\cite{herzig2023incorporating} utilizes scene graphs to fine-tune the pretrained vision-language model.  TSVLC~\cite{doveh2023teaching} leverages language models to generate positive and negative captions. COCO-Counterfactual~\cite{le2023cococounterfactuals} further utilizes both negative images and negative captions to fine-tune CLIP via a rule-based data curation pipeline.

Moreover, previous works mainly consider composition reasoning with semantically different concepts rather than physically grounded relationships such as object counting and distinguishing left/right and above/below. Though a recent work~\cite{paiss2023teaching} improved CLIP's counting capabilities, they neither addressed compositional reasoning nor leveraged negative images. In our paper, we systematically improve both semantic and physical compositional reasoning by leveraging the most capable generative models including GPT-4V~\cite{openai2023GPT4V} and DALLE-3~\cite{openai2023DALLE3} to generate counterfactual image/text pairs.

\subsection{Counterfactual Reasoning}

\begin{figure*}[ht]
\centering
\includegraphics[width=\linewidth]{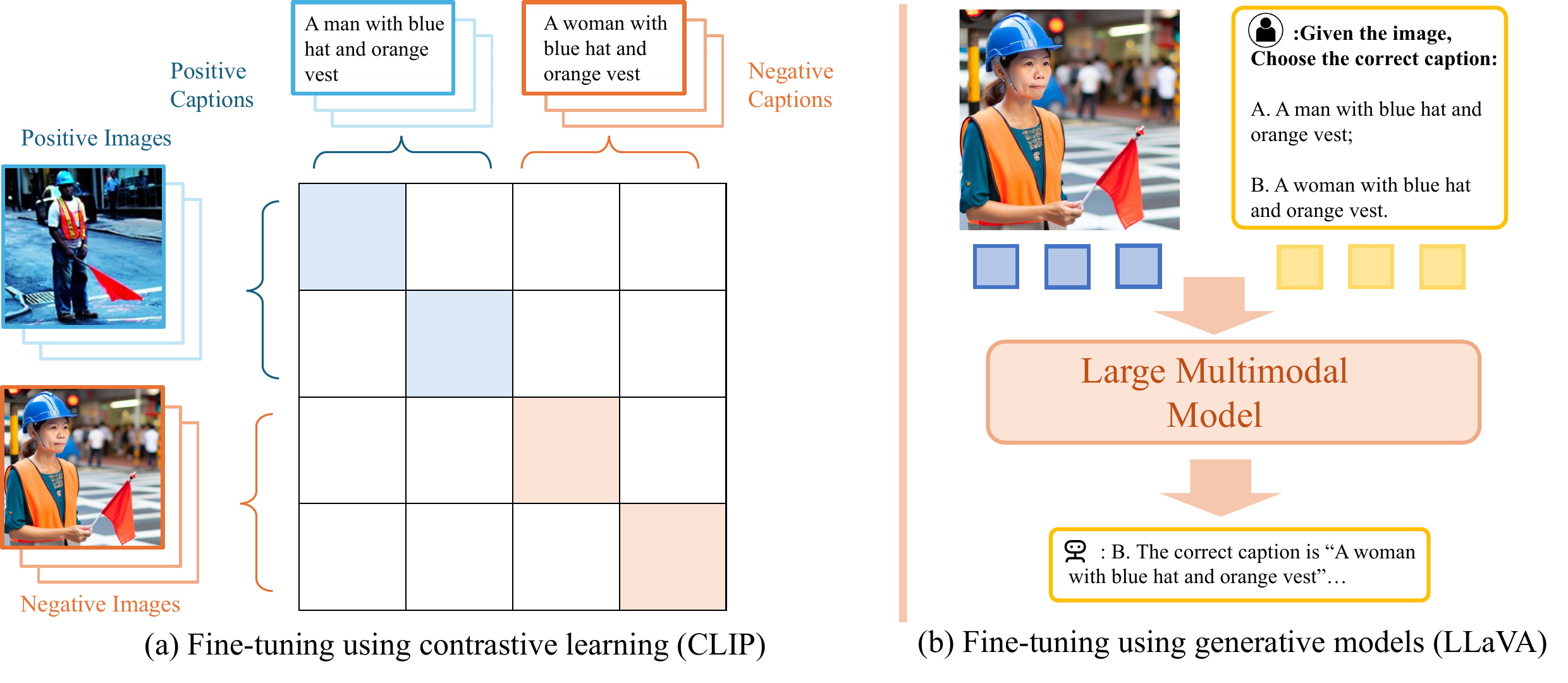}
\caption{Fine-tuning different types of large multimodal models with \newshortname{}. Our pipeline can enhance both contrastive learning models and generative models by augmenting vanilla image-caption pairs with curated negative images and captions. Specifically, our counterfactual image-caption pairs (a) provide auxiliary contrastive loss for models like CLIP, where positive contrastive units in the similarity matrix are colored as blue/red and negative ones are colored as white, and (b) can be naturally integrated into the original next-token prediction loss in text generation models such as LLaVA.
}

\label{fig:trainanymodel}
\end{figure*}

Counterfactual reasoning~\cite{morgan2015counterfactuals} refers to the process of imagining ``what if'' scenarios by modifying the input data. In the context of visio-linguistics scenarios, this involves curating negative images and captions by manipulating the original data and observing how it affects the outcome. This helps models understand cause and effect and predict outcomes in situations they've never seen before. The key to counterfactual reasoning is curating meaningful and hard enough examples. COCO-Counterfactual~\cite{le2023cococounterfactuals} explores simple linguistic rules to generate negative captions and uses an image editing model Prompt2Prompt~\cite{hertz2022prompt} to produce negative images~\cite{hertz2022prompt}. Prompt2Prompt is less desirable in understanding complex language instructions, as the generated counterfactuals are less reliable and challenging. 

In this paper, we collect the very challenging negative image and text examples by leveraging the most capable language models GPT-4~\cite{openai2023gpt4}, text-to-image generation model DALLE-3~\cite{openai2023DALLE3}, and GLIGEN~\cite{li2023gligen}, significantly improving our model's reasoning. Utilizing these hard negatives, our models can better grasp the nuances of language and vision, enhancing their performance on tasks that require a sophisticated understanding of the world.

\section{Approach}

In Sec.~\ref{sec:source data selection}, we first explain our selection of the Flickr30k Entities dataset for both benchmarking and fine-tuning. Then in Sec.~\ref{sec:flickr30k-pos} and \ref{sec:method-counting}, we demonstrate the near-chance performance of multimodal models like CLIP and LLaVA on physically grounded compositional reasoning tasks, including positional understanding and counting, and improve their performance by generating negative image-text pairs via data augmentation and grounded image inpainting. Finally, we introduce using capable text and image generation models to improve models' semantic compositional reasoning in Sec.~\ref{sec:method-attribute}. 

\subsection{Necessity of Grounded Image Captions}
\label{sec:source data selection}

Previous approaches~\cite{yuksekgonul2023when,hsieh2023sugarcrepe} adopt global image captioned datasets such as COCO-Captions~\cite{chen2015microsoft} and craft negative captions to either benchmark or improve compositional reasoning. This method, however, lacks regional information important for physically grounded reasoning, such as determining the relative position of two objects. Therefore, we leverage Flickr30k Entities~\cite{plummer2016flickr30k}, a grounded image captioning dataset to both curate negative image-caption pairs and benchmark models.

\subsection{Improving Positional Understanding}
\label{sec:flickr30k-pos}

Currently, there is no benchmark that directly evaluates the positional understanding of multimodal models. We build such a benchmark, Flickr30k-Positions as in Figure \ref{fig:data_curation} (a), that is composed of positive and negative image-caption pairs describing the same objects in opposite positions.

\vspace{-5pt}
\paragraph{Generating negative captions.}
Specifically, we utilize the bounding boxes $$B=\{B_j=(x_{j1},y_{j1},x_{j2},y_{j2})\mid\forall j\in I\}$$ in Flickr30k Entities~\cite{plummer2016flickr30k} for each object $j$ in Image $I$, where box $B_j$'s upper-left and lower-right corners are at $(x_{j1}, y_{j1})$ and $(x_{j2},y_{j2})$, respectively. In order to ensure non-ambiguity in relative position descriptions between categories (e.g., ``the dog is to the left of the cat''), we only consider images where for each object category, there is only at most one instance of it in the image.  For each pair of objects $\{a,b\}\subseteq I$, we identify the positional object pairs via the formula:

\vspace{-1em}
$$
    \begin{cases}
        x_{a2}\le x_{b1} \implies\text{ $a$ is to the \textit{left} of $b$}\\
        x_{a1}\ge x_{b2} \implies\text{ $a$ is to the \textit{right} of $b$}\\
        y_{a2}\le y_{b1} \implies\text{ $a$ is \textit{above} $b$}\\
        y_{a1}\ge y_{b2} \implies\text{ $a$ is \textit{below} $b$}\\
    \end{cases}
$$

With the formula, we generate positive and negative positional captions $C$ and $C'$ such as $C=$\textit{``a bike is to the left of a woman''} vs. $C'=$\textit{``a bike is to the right of a women''} and $C=$\textit{``a table is above a towel''} vs. $C'=$\textit{``a table is below a towel''}. 

\begin{figure}[t]
    \centering
    \includegraphics[width=\linewidth]{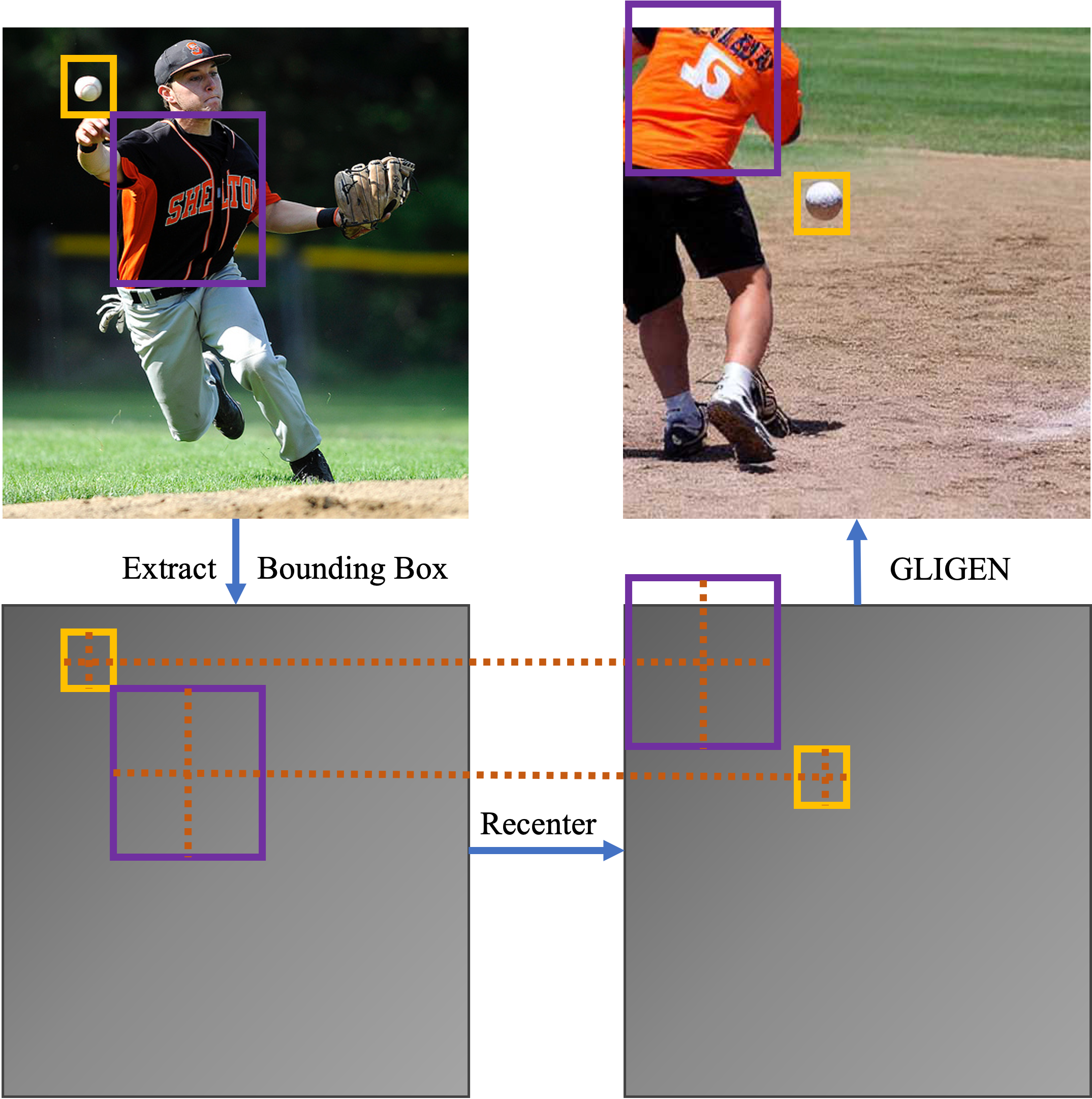}
    \caption{To generate the correct above-below negative image via GLIGEN with the original prompt "\textit{the ball is below the sports outfit}", we recenter the bounding boxes of "\textit{ball}" and "\textit{sports outfit}" and feed them into GLIGEN together with an expanded prompt from GPT4.}
    \label{fig:gligen-gen-ud}
\end{figure}

We also prompt text-only GPT4 \cite{openai2023gpt4} to rewrite the vanilla negative prompt \textit{``$a$ is above/below $b$''} to better guide image generation models. For example, the caption ``a street is above a young child" is unnatural, while the GPT-expanded caption ``the street stretches out above a young child, elevated by a bridge'' is much better. For detailed prompt engineering, see Appendix \ref{sec:pos-gpt4v}. 
\vspace{-5pt}
\paragraph{Generating negative images.}
Furthermore, we include negative images in our dataset to curate Flickr30k-Positions. 
For the left-right pairs, we apply a simple data augmentation method: flip the original images horizontally. For the above-below pairs, however, flipping vertically in most cases would lead to unreasonable images. To address this issue, we utilize GLIGEN \cite{li2023gligen}, an object-level text-to-image generation model that can ground text prompts with bounding boxes. In particular, for each above-below object pair $(a,b)\in I$, we swap the centers of the bounding boxes $B_a,B_b$ while maintaining their widths and heights, ensuring that the generated images are natural while preserving the objects' sizes and aspect-ratios as demonstrated in Figure \ref{fig:gligen-gen-ud}. The new bounding boxes and prompts are fed to GLIGEN to generate the corresponding negative image $I'$.

\subsection{Improving Object Counting}
\label{sec:method-counting}

We introduce Flickr30k-Counting to enhance LMMs' object counting capabilities. We again curate counterfactual image-caption pairs as in Figure \ref{fig:data_curation} (b). We count the number  of bounding boxes for each object category in Flickr30k-Entities \cite{plummer2016flickr30k}. For two distinct object categories $\{P,Q\}\subseteq I$, we generate a positive caption $C=$\textit{``there are $n_P$ $P$'s and $n_Q$ $Q$'s''}, such as \textit{``there are three cats and four dogs''} where category $P$ is ``cat'' and category $Q$ is ``dog''.

\vspace{-5pt}
\paragraph{Generating negative captions.}

If all instances of the two selected object categories do not spatially overlap with any other object in the image, we generate the corresponding negative caption by decrementing the one with more objects, $Q$, and incrementing the one with fewer, $P$; for example, $C'=$\textit{``there are four cats and three dogs''}. This enforces a hard counterfactual format as the numbers in the positives are reversed in the negatives.

In most cases, however, objects in an image do overlap, and we cannot ascertain that the increment/decrement rule will still generate hard counterfactuals. For example, if a dog overlaps with a cat, and we re-use the rule above to remove one dog and add one cat, we end up with ``three cats and three dogs''. In this case, we only remove either one of $P$ or one of $Q$, for example, the dog, and all objects it overlaps with. An example of the resulting negative caption is $C'=$\textit{``there are two cats and three dogs.''}

\vspace{-5pt}
\paragraph{Generating negative images.}

In the case where objects do not overlap, we simply leverage GLIGEN \cite{li2023gligen} to inpaint a new object of the incremented category $P$ at the bounding box of one of the objects from the decremented category $Q$. Otherwise, we remove the target objects using GLIGEN to replace it with an object from a generic category. We simply use the \texttt{plant} category for replacement, as plants can appear in almost any natural setting, let it be outdoor or indoor, and we find that GLIGEN does well in generating it according to each image $I$'s background during inpainting for each corresponding negative image $I'$.
More details can be found in Appendix \ref{sec:objrm-vs-plants}.

\subsection{Improving Semantic Compositional Reasoning}
\label{sec:method-attribute}

Existing works~\cite{yuksekgonul2023when,le2023cococounterfactuals} use a rule-based approach to generate negative images/captions for semantic compositional reasoning, resulting in less desirable negatives such as uncommon negative captions e.g., \textit{``A dog with blue fur"} from \textit{``A dog with black fur"}. 
We introduce Flickr30k-Attributes as in Figure~\ref{fig:data_curation} (c), which utilizes GPT-4V \cite{openai2023GPT4V} to generate hard negative captions and leverages DALLE-3~\cite{openai2023DALLE3} for generating corresponding high-quality negative images. We extract the first and most detailed caption $C$ for each image $I$ from the original Flickr30k dataset \cite{young2014flickr30k}. 
\vspace{-5pt}
\paragraph{Generating negative captions.}

To generate high-quality negative captions, we leverage GPT-4V and feed both positive images $I$ and captions $C$ to generate negative captions $C'$.  To let GPT-4V know the objects' locations, we overlay bounding boxes $B$ onto the original positive image~\cite{cai2023vipllava} and also feed this annotated image into GPT-4V.
We prompt GPT-4V to generate three kinds of captions, including (i) changing a noun, (ii) changing an adjective, and (iii) swapping any of the two phrases, nouns, or adjectives. For example, given the original caption $C$ \textit{``a man wearing a black shirt and blue jeans"}, GPT-4V can generate the following negatives $C'$: (i) \textit{``a man wearing a black \textbf{jacket} and blue jeans"}, (ii) \textit{``a man wearing a black shirt and \textbf{red} jeans"}, and (iii) \textit{``a man wearing a \textbf{blue} shirt and \textbf{black} jeans"}. Detailed prompts to GPT-4V are shown in Appendix \ref{sec:attr-gpt4v}.

\vspace{-5pt}
\paragraph{Generating negative images.}
We simply feed the curated captions $C'$ from GPT-4V directly into the currently most capable text-to-image generation model, DALLE-3~\cite{openai2023DALLE3}, to generate high-quality negative images $I'$.

\subsection{Fine-tuning Methodology}

As demonstrated in Figure~\ref{fig:trainanymodel}, \newshortname{} can be used to fine-tune both contrastive~\cite{radford2021learning} and generative~\cite{liu2023llava} multimodal models. To fine-tune a contrastive model like CLIP, we sample $N$ positive image-text pairs and the corresponding negative image-text pairs to form a batch $(C,I,C',I')$ as shown in Figure~\ref{fig:trainanymodel} (a). We refer to this method as \textbf{Grouping}, where the model is forced to distinguish the positive pairs from their corresponding negatives, which is demonstrated to be helpful in our experiments. We also conduct a detailed analysis of the effectiveness of this method in Section \ref{para:ablation}. The training loss is the same as the original, where cross-entropy loss is used to maximize the diagonal elements and minimize all other entries in the similarity matrix.

For generative models like LLaVA~\cite{liu2023llava,liu2023improved,liu2024llavanext}, we reformat our data into conversations and follow their exact training paradigm to fine-tune, as shown in Figure~\ref{fig:trainanymodel} (b). The training loss is the original next-token prediction loss. 
\section{Experiments}

\begin{figure}[t]
    \centering
    \includegraphics[width=\linewidth]{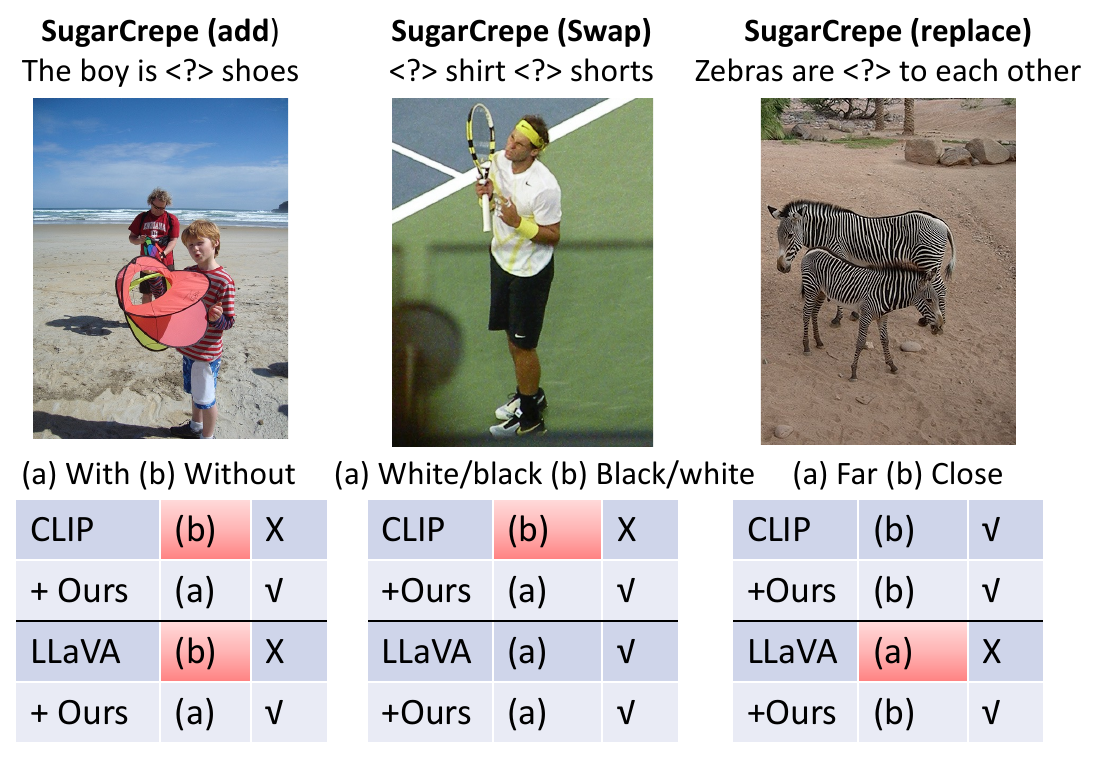}
    \caption{Qualitative examples of models' compositional reasoning capabilities before/after being finetuned via our approach \newshortname{}. Wrong answers are marked in red. Our approach enhances both CLIP and LLaVA's reasoning capabilities.}
    \label{fig:examples}
\end{figure}

We utilize the proposed three datasets, Flickr30k-Positions, Flickr30k-Counting, and Flickr30k-Attributes, created via \newshortname{}, to fine-tune both contrastive and generative models. Specifically, we select two common multimodal models, ViT-B/32 from~\cite{ilharco2021openclip} OpenCLIP pre-trained on LAION-2B~\cite{schuhmann2022laion5b} and LLaVA-1.5~\cite{liu2023improved} as base models. An overview of our results is shown in Figure \ref{fig:examples}.

\subsection{Implementation Details}

When generating images for Flickr30k-Attributes, we provide DALLE-3 \cite{openai2023DALLE3} with \texttt{hd} quality and \texttt{natural} style.
We train CLIP~\cite{radford2021learning,ilharco2021openclip} and LLaVA \cite{liu2023improved} on 1 and 4 NVIDIA A100 80GB GPUs, respectively. When fine-tuning CLIP, we set Adam optimizer's $\beta_1$ and $\beta_2$ to be 0.9 and 0.98, and weight decay to be 0.2.

For Flickr30k-Attributes, we train the CLIP model with a learning rate of 1e-5, batch size of 256, and mixed precision for 5 epochs. For LLaVA-1.5, we use a learning rate of 2e-6 and batch size of 16 for 1 epoch. For Flickr30k-Positions, we use a learning rate of 2.56e-5 and batch size of 1024 for CLIP, and train for 50 epochs without grouping and 15 epochs with grouping. For Flickr30k-Counting, we train CLIP with a learning rate of 5e-5. The LLaVA parameters for both Flickr30k-Positions and Flickr30k-Counting are identical, with a learning rate of 2.56e-5 and batch size of 16.

\subsection{Evaluating Positional Understanding}
\label{sec:test-pos}

To the best of our knowledge, we are the first to comprehensively evaluate a multimodal model's ability in understanding the left-and-right and/or above-and-below positioning relations between objects. We conduct a 4:1 train-test split on Flickr30k-Positions and use the test subset as the benchmark. We further separate this dataset into 3 subsets: left-and-right only, above-and-below only, and both. We evaluate our fine-tuned CLIP models by comparing the CLIP similarity scores between the four image-caption pairs, $(C, I), (C, I'), (C', I), (C', I')$ from each group of data in the dataset, where $(C, I, C', I')$ denotes the positive and negative image/caption, respectively. The model receives a score of 0.5 if $s_{(C,I)}>s_{(C',I)}$ and another score of 0.5 if $s_{(C,I')}<s_{(C',I')}$, where $s_{(C,I)}$ is the CLIP cosine-similarity score between caption $C$ and image $I$. For LLaVA-1.5, we simply query the model to choose between the positive and negative captions when presented with the ground truth image. The results are shown in Table \ref{tab:pos-comparison}.

\begin{table}[t]
\centering
\small
\begin{tabular}{l|ccc}
\toprule
Models & LR & AB & Both\\
\midrule
Random & 50.00 & 50.00 & 50.00\\
\midrule
CLIP & 50.55 & 52.80 & 51.56\\
+ \newshortname{} & \textbf{75.88} & \textbf{91.52} & \textbf{84.90}\\
\midrule
LLaVA-1.5 & 61.84 & 56.69 & 59.17\\
+ \newshortname{} & \textbf{95.72} & \textbf{96.21} & \textbf{96.00}\\
\bottomrule
\end{tabular}
\caption{Performance of CLIP and LLaVA on Flickr30k-Positions's test dataset. ``LR'' means left-and-right, and ``AB'' means above-and-below.}
\label{tab:pos-comparison}
\end{table}

As we hypothesized, LMMs are indeed largely oblivious to the objects' positioning in the image, which is especially manifested in the vanilla CLIP's performance, which is only marginally better than random guessing. Vanilla LLaVA-1.5 shows only slightly better performance.

After fine-tuning with the training split of Flickr30k-Positions, both models perform significantly better across all subsets. Specifically, for the mixed case, CLIP improves by 33\% points, and LLaVA achieves a high accuracy of 96\%. These results demonstrate that \newshortname{} is highly effective across different kinds of multimodal models.

\vspace{-0.3em}
\subsection{Evaluating Object Counting}
\label{sec:exp-counting}
\vspace{-0.3em}

Here we show the counting capability of our models fine-tuned on Flickr30k-Counting. We select a different dataset, PointQA-LookTwice \cite{mani2022point}, as the evaluation benchmark. Specifically, PointQA-LookTwice was designed to ask models the number $n_J$ of occurrences of an object category $J$ in an image $I$. 
We reformat this dataset such that for every $J$, we generate a positive caption $C_J=$ \textit{``There are $n_J$ $J$'s''} and a negative caption $C_J'=$ \textit{``There are $n_J+1$ $J$'s''}. For example, given $C_J$ \textit{``there are 3 dogs''}, $C_J'$ is \textit{``there are 4 dogs''}.
Similar to what we did in Section~\ref{sec:test-pos}, we mark that a model made a correct prediction if (i) CLIP shows $s_{(C_J,I)}>s_{(C_J',I)}$ and (ii) LLaVA-1.5 correctly chooses the option for $C_J$. The results are shown in Table~\ref{tab:counting-comparison}.

\begin{table}[t]
\centering
\small
\begin{tabular}{l|cc}
\toprule
 & CLIP & LLaVA-1.5 \\
\midrule
Vanilla & 57.50 & 44.87 \\
\midrule
+ \newshortname{} & \textbf{68.51} & \textbf{50.74}\\
\bottomrule
\end{tabular}
\caption{Comparison between CLIP and LLaVA on the benchmark created out of PointQA-LookTwice.}
\label{tab:counting-comparison}
\end{table}

As CLIP performs slightly better than random guessing, it is surprising that LLaVA-1.5 performs worse than random. Nevertheless, fine-tuning with Flickr30k-Counting improves both models' counting capability. This shows the effectiveness of using GLIGEN-generated negative images in \newshortname{} to tackle the problem of counting. We conduct further ablation studies in Appendix \ref{sec:counting-more-exp}.

\begin{table*}[hbtp]
\centering
\small
\begin{tabular}{l|ccl|ccl}
\toprule
Categories & CLIP & NegCLIP & + \newshortname{} &   LLaVA-1.5 & GPT-4V  &+ \newshortname{}  \\
\midrule
Add  & 80.22 & 87.28 & \textbf{89.44} \textcolor{teal}{(+4.62)} & 86.02 & 91.63 & \textbf{97.13} \textcolor{teal}{(+11.11)}\\
Replace  & 82.40 & 85.36 & \textbf{87.10} \textcolor{teal}{(+4.70)} & 92.38 & \textbf{93.53} & 92.82 \textcolor{teal}{(+0.43)}\\
Swap  & 65.42 & \textbf{75.31} & 72.22 \textcolor{teal}{(+6.80)} & 85.95 & 88.21 & \textbf{90.88} \textcolor{teal}{(+4.93)}\\
\midrule
Average & 79.54 & 84.85 & \textbf{86.15} \textcolor{teal}{(+6.61)} & 89.27 & 92.19 & \textbf{94.17} \textcolor{teal}{(+4.90)} \\
\bottomrule
\end{tabular}
\caption{Comparison between performances of CLIP and LLaVA-1.5 on SugarCrepe before and after fine-tuning. We also add NegCLIP~\cite{yuksekgonul2023when} and GPT-4V~\cite{openai2023GPT4V} as strong baselines for contrastive and generative multimodal models. The best performances are bolded, and improvements against CLIP and LLaVA are measured in the parentheses. \newshortname{} shows significant performance boost compared to both vanilla CLIP/LLaVA-1.5 model and advanced models such as GPT-4V.}
\label{tab:sugarcrepe-comparison}
\end{table*}

\subsection{Evaluating Semantic Compositional Reasoning}
\label{sec:test-attr}

Similar to the setup in Sec~\ref{sec:exp-counting}, we fine-tune both CLIP and LLaVA-1.5 under our curated Flickr30k-Attributes.  We use another common benchmark SugarCrepe \cite{hsieh2023sugarcrepe} to evaluate the model's semantic compositional reasoning capabilities. SugarCrepe has three major categories depicted in Table \ref{tab:sugarcrepe-comparison}. The evaluation protocol is also the same as Sec~\ref{sec:exp-counting}.  
We choose NegCLIP \cite{yuksekgonul2023when} and GPT-4V \cite{openai2023GPT4V} as the representative strong baselines for contrastive and generative models respectively.

We observe significant improvements for both CLIP and LLaVA-1.5, both on average and categorically. Summarized results are shown in Table~\ref{tab:sugarcrepe-comparison}, where Appendix \ref{sec:sugarcrepe-comparison-extend} contains more detailed scores.

For example, \newshortname{} fine-tuned CLIP surpasses NegCLIP on average as well as in two main categories. Note that the source of fine-tuning data, Flickr30k-Entities, contains much fewer image-caption pairs than COCO-Captions (around 100k image-caption pairs) where NegCLIP is trained.
Furthermore, when prompting GPT-4V to generate the negative captions, we intentionally prompt GPT-4V to produce \texttt{None} when there is no feasible condition to conduct ``swapping". These two factors lead to our data curation pipeline resulting in much fewer negative samples for the ``swap'' category, which is around 2.5k. We argue that using our pipeline on other datasets that can generate more of the ``swap'' attribute should lead to larger improvements in performance.

LLaVA-1.5 also performs better in all three categories after fine-tuning. It is also surprising that our fine-tuned model outperforms the SoTA LMM GPT-4V both on average and in two of the categories, most significantly for the ``add'' category. We observe improvements on other datasets~\cite{diwan2022winoground} as well in Appendix \ref{sec:test-wino}. The promising results of LLaVA, such as the score of 94.17\% on SugarCrepe surpassing GPT-4V and a 10\% improvement on Flickr30k-Position's above-below subset compared to GPT-4V, showcases that \newshortname{} effectively improves LLaVA's semantic and physical reasoning capabilities, and we thus identify \newshortname{} as a valid candidate towards improving SoTA LMMs such as GPT-4V as well.

\newshortname{} shows better performance compared to the prior rule-based methods or less desirable models that generates negatives. These significant improvements show that fine-tuning with accurate and hard negative samples is important, again demonstrating the effectiveness of our data curation and fine-tuning pipeline for both contrastive models and generative models for improving semantic counterfactual understanding.

\subsection{In-Depth Analysis}

\paragraph{Effectiveness of Negative Images, Negative Captions, and Grouping}
\label{para:ablation}
The core of \newshortname{} is to (i) fine-tune models with both negative images and negative captions and (ii) use grouping to help the model better distinguish the positive pairs from the negatives. Here we conduct rigorous ablation studies to demonstrate the necessity of each component. We use the same training parameters and test with the same methods as we did for fine-tuning CLIP and evaluating it on Flickr30k-Positions.

\begin{table}[t]
\centering
\small
\begin{tabular}{ccc|ccc}
\toprule
\begin{tabular}[c]{@{}c@{}}Negative\\ Images\end{tabular} & \begin{tabular}[c]{@{}c@{}}Negative\\ Captions\end{tabular} & \begin{tabular}[c]{@{}c@{}}Group\\-ing\end{tabular} & LR & AB & Both\\
\midrule
$\times$ & $\times$ & $\times$ & 50.55 & 52.80 & 51.56\\
\midrule
$\times$ & $\checkmark$ & $\times$ & 55.86 & 62.25 & 58.68\\
$\checkmark$ & $\times$ & $\times$ & 54.35 & 56.79 & 54.95\\
\midrule
$\checkmark$ & $\checkmark$ & $\times$ & 69.99 & 91.24 & 76.88\\
$\checkmark$ & $\checkmark$ & $\checkmark$ & \textbf{75.88} & \textbf{91.52} & \textbf{84.90}\\
\bottomrule
\end{tabular}
\caption{Ablation study demonstrating the necessity of using (i) negative images, (ii) negative captions, and (iii) grouping strategies. Models are fine-tuned and evaluated on the Flickr30k-Positions dataset.}
\label{tab:ablation}
\end{table}

As shown in Table~\ref{tab:ablation}, even though using either the negative caption or the negative image can improve performance, the scores are significantly improved when using both elements to fine-tune. This demonstrates that \newshortname{}, which incorporates both negative captions and negative images, is necessary to achieve desirable improvements. Grouping also delivers further improvements compared to not using this strategy.

\paragraph{Correctness of DALLE-3 generated images}

We randomly sample 300 DALLE-3 images generated from the negative captions in Flickr30k-Attributes for human annotators to check whether the generated image is consistent (matches) with the negative captions. We obtain a consistency score of 84.67\%, which demonstrates the high quality of the DALLE-3 generated images.

\begin{table*}[t]
\centering
\small
\begin{tabular}{l|llllllll}
\toprule
\textbf{}        & MMBench & {MMBench\_cn} & {LLaVA\_W} & POPE & {ScienceQA} & {MM-Vet} & {VizWiz} & {MME}  \\
\midrule
LLaVA-1.5        & 67.7             & 63.6                 & 70.7 & \textbf{85.9}              & 71.6               & 35.4            & 53.6            & \textbf{1531} \\
+ \newshortname{} & \textbf{68.6}    & \textbf{64.1}        & \textbf{71.1}   & 85.6          & \textbf{72.1}      & \textbf{39.5}   & \textbf{54.2}   & 1510\\         
\bottomrule
\end{tabular}
\caption{Evaluating our fine-tuned LLaVA-1.5 on its original benchmarks.}
\label{tab:down-llava}
\end{table*}

\begin{table}[t]
\centering
\small
\begin{tabular}{l|cc}
\toprule
Scores & CLIP & + \newshortname{} \\
\midrule
Image @1 & 39.44 & 37.81\\
Image @5 & 65.43 & 64.24\\
\midrule
Text @1 & 56.48 & 56.96\\
Text @5 & 79.74 & 80.12\\
\midrule
Average @1 & 47.96 & 47.39\\
Average @5 & 72.59 & 72.18\\
\bottomrule
\end{tabular}
\caption{Comparison of image and text retrieval precision scores on the MSCOCO dataset between the original CLIP model and \newshortname{} fine-tuned CLIP. The latter is able to maintain overall performance with minor improvements in text retrieval precision.}
\label{tab:retrieval}
\end{table}

\paragraph{Performance on zero-shot vision-language tasks}
We evaluate whether fine-tuning on the counterfactual data hurts the original zero-shot vision-language performance. For CLIP, we compare the image and text retrieval performance on MSCOCO \cite{lin2015microsoft} of the original CLIP model and \newshortname{} fine-tuned model on Flickr30k-Attributes. The results in Table \ref{tab:retrieval} show that on average, the model fine-tuned via \newshortname{} maintains performance with marginal difference compared to the original CLIP model. In Appendix~\ref{sec:down-clip}, we also show that fine-tuning with \newshortname{} significantly improves CLIP's performance on several common downstream tasks. For LLaVA, we mix Flickr30k-Attributes with LLaVA's original training data before fine-tuning our model. Compared with vanilla LLaVA-1.5 in Table~\ref{tab:down-llava} on the same benchmarks from \cite{liu2023improved}, the \newshortname{}-finetuned LLaVA overall performs similarly (or even slightly better). From these studies, we see that improving LMMs' semantical compositional reasoning with Flickr30k-Attributes does not hurt downstream performance of both contrastive and generative models.

\section{Conclusion}

In conclusion, \newshortname{} significantly enhances the visio-linguistic compositional reasoning capabilities of multimodal contrastive and generative models. This is achieved by addressing the neglect of physically grounded reasoning and exploiting the potential of using text and image generation models for semantic counterfactual fine-tuning. We believe our contributions can pave the way for further research in compositional reasoning.

\section*{Acknowledgement}
This work was supported in part by NSF CAREER IIS2150012, Sony Focused Research award, and Institute of Information \& communications Technology Planning \& Evaluation(IITP) grants funded by the Korea government(MSIT) (No. 2022-0-00871, Development of AI Autonomy and Knowledge Enhancement for AI Agent Collaboration) and (No. RS2022-00187238, Development of Large Korean Language Model Technology for Efficient Pre-training), and Microsoft Accelerate Foundation Models Research Program.


This research is funded in part by the University of Wisconsin–Madison L\&S Honors Program through a Trewartha Senior Thesis Research Grant.

\section*{Limitations}
We acknowledge that DALLE-3 and GPT4-V are closed-source models, which makes it difficult to systematically analyze their behavior.

\bibliography{custom}

\appendix
\section*{Appendix}
\label{sec:appendix}

\begin{table*}[hbtp]
\centering
\small
\begin{tabular}{l|ccl|ccl}
\toprule
Categories & CLIP & NegCLIP & + \newshortname{} &   LLaVA-1.5 & GPT-4V  &+ \newshortname{}  \\
\midrule
Add Object & 77.89 & 88.78 & \textbf{90.35} \textcolor{teal}{(+3.20)} & 87.83 & 91.59 & \textbf{97.82} \textcolor{teal}{(+9.99)}\\
Add Attribute & 87.15 & 82.80 & \textbf{86.71} \textcolor{teal}{(+8.82)} & 80.64 & 91.76 &\textbf{95.09} \textcolor{teal}{(+14.45)}\\
Replace Object & 93.77 & 92.68 & \textbf{95.94} \textcolor{teal}{(+2.17)}  & 96.73 & 96.31 & \textbf{98.37} \textcolor{teal}{(+1.64)}\\
Replace Attribute & 82.61 & 85.91 & \textbf{87.94} \textcolor{teal}{(+5.33)} &	 92.64 & \textbf{93.53} & \textbf{93.53}   \textcolor{teal}{(+0.89)}\\
Replace Relations & 68.92 & \textbf{76.46} & \textbf{76.24} \textcolor{teal}{(+7.32)}  & {87.13} &\textbf{90.26} &  85.92 \textcolor{gray}{(-1.21)}\\
Swap Object & 60.00 & \textbf{75.51} & 68.57 \textcolor{teal}{(+8.57)} &  84.90 &  83.13& \textbf{85.72} \textcolor{teal}{(+0.82)}\\
Swap Attribute & 67.42 & \textbf{75.23} & 73.57 \textcolor{teal}{(+6.15)}  & 86.34 & 90.09 & \textbf{92.79} \textcolor{teal}{(+6.45)}\\
\midrule
Average & 79.54 & 84.85 & \textbf{86.15} \textcolor{teal}{(+6.61)} & 89.27 & 92.19 & \textbf{94.17} \textcolor{teal}{(+4.90)} \\
\bottomrule
\end{tabular}
\caption{Detailed version of Table \ref{tab:sugarcrepe-comparison}: comparison of performance over each sub-category.}
\vspace{-1em}
\label{tab:sugarcrepe-comparison-extend}
\end{table*}

\section{Prompt guidance to GPT-4V for Flickr30k-Attributes}
\label{sec:attr-gpt4v}
As a minor detail specific to the dataset itself, we extracted the objectified caption $C_I'$ from the annotated Flickr30k-Entities dataset \cite{plummer2016flickr30k}. The objectified captions have square brackets added to the original caption that allow us and the LMMs to refer to each phrase for the object $j$ in the caption with a unique ID $\#_j$.

The exact prompt guidance we provided to GPT-4V \cite{openai2023GPT4V} is as follows:

You are given an image, the same image but with bounding boxes, its corresponding caption and an enhanced form of the caption. Their format is as follows:
Original Caption: A child in a pink dress is helping a baby in a blue dress climb up a set of stairs in an entry way.
Enhanced Caption: [/EN\#1/people A child] in [/EN\#2/clothing a pink dress] helping  [/EN\#3/people a baby] in  [/EN\#4/clothing a blue dress] climb up [/EN\#5/other a set of stairs] in [/EN\#6/scene an entry way].
In the enhanced caption, there is no new data, but that each “entity” is marked by a pair of square brackets. Most entities each correspond to one or more bounding boxes, which will be specified. For example, entity 1 in the sentence is “A child”, which is marked by a tag [/EN\#1/people …]. “people” states the type of the entity. If entity is “other”, then there are no restrictions applied.

You are tasked to:

Generate a caption that changes the object being discussed in exactly one of the entities. You MUST ensure that the new object is the same type of entity as the original object as specified in the tag. For example: [/EN\#1/people A child] => [/EN\#1/people An adult] is allowed, but [/EN\#1/people A child] => [/EN\#1/people A cat] is not allowed because a cat is not “people”;

Generate a caption that changes the qualifier (such as an adjective of a quantifier) that describes the object in exactly one of the entities. For example: [/EN\#2/clothing a pink dress] => [/EN\#2/clothing a green dress].

Generate, if possible, a caption that reverses two of the entities or their qualifiers such that the original sentence structure is not changed, but produces a negative prompt. For example, given two entities “a green dress” and “a blue blouse”, you can either swap the two entities’ order or swap the adjectives and produce “a blue dress” and “a green blouse”. If you cannot generate one, report None.
                    
All in all, the new description must meet all of these requirements:

1. The change of attribute must be sufficiently different to make the new description inaccurate, but it should also be somewhat related to be challenging to an AI model.

2. Compared to the original description, the new description must differ in only one attribute. All other details must be kept the same.

3. The new description must mimic the sentence structure of the original description.

4. The new description must be fluent, logical, and grammatically correct.

5. Carefully look at the image, and give negative captions that are reasonable given the objects’ position, size, and relationship to the overall setting.

6. Pose challenging(difficult enough) negative captions so that a large multimodal text generation model should struggle to distinguish the original caption v.s. negative caption.

Here are some examples whose output format you should follow:
Original Caption: A child in a pink dress is helping a baby in a blue dress climb up a set of stairs in an entry way.
Enhanced Caption: [/EN\#1/people A child] in [/EN\#2/clothing a pink blouse] helping  [/EN\#3/people a baby] in  [/EN\#4/clothing a blue dress] climb up [/EN\#5/other a set of stairs] in [/EN\#6/scene an entry way].
Bounding Boxes: \#1: purple
Your answer:
\{“noun”: \{“action”: (1, “a child”, “an adult”), “caption”: “An adult in a green dress is helping a baby in a blue dress climb up a set of stairs in an entry way.”]\}, “adjective”: \{“action”: (2, “a pink dress”, “a green dress”), “caption”: “A child in a green dress is helping a baby in a blue dress climb up a set of stairs in an entry way.”\}, “reverse”: \{“action”: (2, 4), “caption”: “A child in a blue blouse is helping a baby in a pink dress climb up a set of stairs in an entry way.”\}\}

\section{Prompt guidance to GPT-4V for Flickr30k-Positions (above-and-below)}
\label{sec:pos-gpt4v}
To generate high-quality counterfactual images for the above-and-below subset of Flickr30k-Positions, we use the re-writing technique to generate convincing and detailed captions for GLIGEN \cite{li2023gligen}. Specifically, we leverage GPT-4V \cite{openai2023GPT4V} with the following prompt:

I will give you a caption in the format "A is above B." You need to expand the sentence such that the meaning "A is above B" is preserved and your answer is reasonable for a human to understand what you're describing. Do not make the answer too long; one long sentence is enough. For example, if i give you "a man is under a dog", a good answer would be "there is a man resting on the ground, and there is a dog lying above him." One restriction: A and B do not overlap. This means that if I ask you to expand "a hat is below water", you must not assume that the hat is below water. Remember that you MUST include both A and B in your answer, like my example did.

\section{Object Removal vs Inpainting Plants}
\label{sec:objrm-vs-plants}
\begin{figure}[hbtp]
\centering
\begin{subfigure}[b]{\linewidth}
\centering
\includegraphics[width=\linewidth]{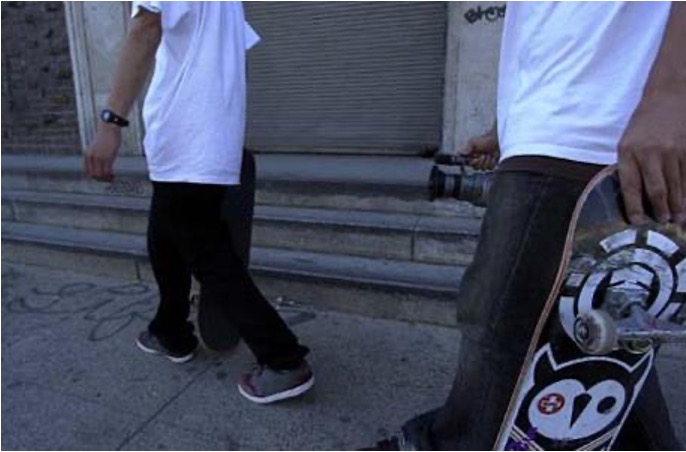}
\caption{Original Image}
\end{subfigure}
\begin{subfigure}[b]{0.49\linewidth}
\centering
\includegraphics[width=\linewidth, height=3cm]{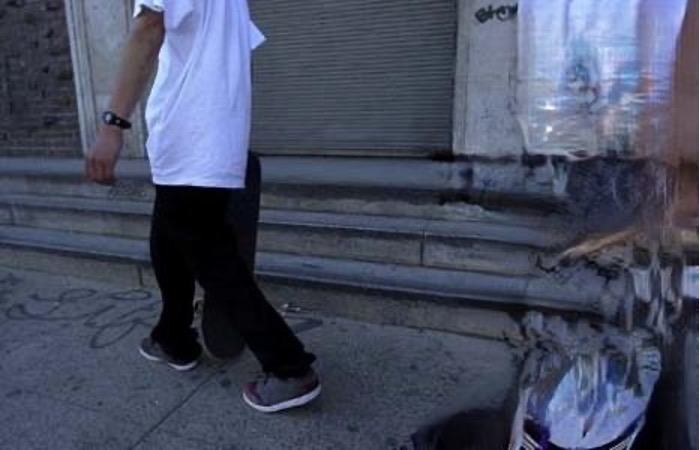}
\caption{Object Removal}
\end{subfigure}
\begin{subfigure}[b]{0.49\linewidth}
\centering
\includegraphics[width=\linewidth, height=3cm]{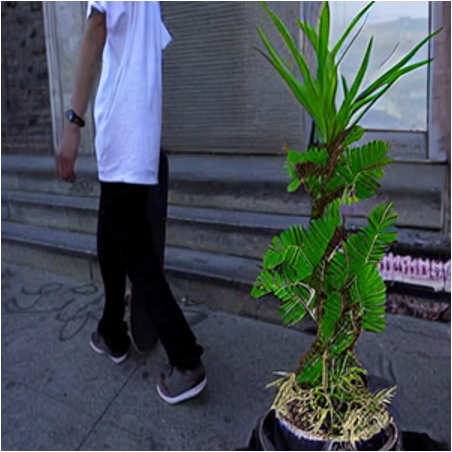}
\caption{GLIGEN Inpainted Plant}
\end{subfigure}
\caption{We want to remove/cover the person on the right with their skateboard in image (a) completely. Object removal only achieves the goal partially and makes the image much less natural, while GLIGEN inpainted a plant perfectly into the image as it also preserved the rest of the information in the image.}
\label{fig:remove-vs-inpaint}
\end{figure}

We explain here as for why we do not use object removal tools to curate our data for Flickr30k-Counting as we make one example in Figure \ref{fig:remove-vs-inpaint}. We used a tool from a public Github Repository 
\url{https://github.com/treeebooor/object-remove} and compared it with GLIGEN \cite{li2023gligen}'s inpainting results. We found a significant pattern that when object removal failed, most of GLIGEN's inpainting succeeded. 

\section{Overfitting in Above-Below Subset?}
In section \ref{sec:test-pos}, we observed that both models performed better on the above-and-below subset, with CLIP~\cite{radford2021learning, ilharco2021openclip} performing significantly better. However, we argue that this is not caused by overfitting because there is an even number of both "A is above B" captions vs. "A is below B" captions, and both captions are not always matched to a generated image by GLIGEN \cite{li2023gligen} or the original image from Flickr30k \cite{young2014flickr30k}. Since the training and testing datasets are shuffled and separated from the same dataset, this ensures that neither model can find a pattern in, for example, recognizing the generated image apart from the original image and choosing an option on that basis. Another proof is that since LLaVA-1.5 \cite{liu2023improved} can perform equally well on both subsets after fine-tuning, the two subsets are shown to be at least as well-defined as the other. We also welcome others to use our dataset as a benchmark to test their model's ability to understand object positioning in images.

\section{Results on the Winoground Dataset}
\label{sec:test-wino}
We evaluated our LLaVA-1.5 \cite{liu2023improved} on Winoground \cite{diwan2022winoground}, a model specifically made to test a model's limit of visio-linguistic capabilities via human-annoted difficult image-caption pairing questions. Our model, after fine-tuning on Flickr30k-Attributes, showed significant improvements on the text score of Winoground.
\begin{table}[htbp]
    \centering
    \label{tab:my_label}
    \small
    \begin{tabular}{ll}
        \toprule
        Model &  Text Score \\
        \midrule
        CLIP$_{\text{(ViT-B/32)}}$~\cite{radford2021learning} & 25.25 \\
        UNITER$_{\text{base}}$~\cite{chen2020uniter} & 32.25\\
        UNITER$_{\text{large}}$~\cite{chen2020uniter} & 38.00\\
        VinVL~\cite{zhang2021vinvl} & 37.75\\
         BLIP2~\cite{zhang2021vinvl} & 44.00\\
         PALI~\cite{chen2023pali} & 46.50\\
        \midrule
        LLaVA-1.5~\cite{liu2023improved} & 65.85\\
        LLaVA-1.5 + \newshortname{} & \textbf{69.15} (\textcolor{teal}{+3.30})\\
        \bottomrule
    \end{tabular}
    \caption{Our fine-tuned model of LLaVA-1.5 with Flickr30k-Attributes shows significant improvements on the difficult visio-linguistic reasoning dataset Winoground.}
\end{table}

\section{More Ablations on Flickr30k-Counting}
We also test not using any of the generated negative image-caption pairs and not using grouping while fine-tuning CLIP with Flickr30k-Counting. Evaluation results on PointQA-LookTwice are shown in Table~\ref{tab:counting-comparison-appendix}. Both ablations showed much less improvement, showcasing that using grouping and the negative image-caption pairs is more powerful in improving models' counting abilities.
\label{sec:counting-more-exp}
\begin{table}[hbtp]
\centering
\small
\begin{tabular}{l|l}
\toprule
 Model Setting & Accuracy  \\
\midrule
Vanilla & 57.50  \\
\midrule
+ \newshortname{} (No Neg) & 60.85\\
+ \newshortname{} (No Group) & 65.70 \\
\midrule
+ \newshortname{} & \textbf{68.51}\\
\bottomrule
\end{tabular}
\caption{More ablations on the CLIP model trained with Flickr30k-Counting. the score without any negatives and the score without grouping.}
\label{tab:counting-comparison-appendix}
\end{table}

\section{More Results on SugarCrepe}
\label{sec:sugarcrepe-comparison-extend}
Here we show the performance improvements for every sub-category of SugarCrepe in Table~\ref{tab:sugarcrepe-comparison-extend}. Overall, \newshortname{} shows clear performance gain over the two base models,  CLIP~\cite{radford2021learning} and LLaVA~\cite{liu2023llava}. We also point out that the ``swap object'' category in SugarCrepe \cite{hsieh2023sugarcrepe} only has 246 items as compared to other categories' 500+, 1000+ items; this means that the performance on this specific category could show more fluctuations caused by the training process.

\section{Ratio of Augmentation}
We study the level of effectiveness our dataset Flickr30k-Attributes brings upon the models if we choose to reduce the amount of curated data used during fine-tuning. As we observe in Table~\ref{tab:ratio-llava} and especially Table~\ref{tab:ratio-clip}, using 100\% of our dataset yields the greatest amount of improvements in LLaVA and CLIP's semantical compositional reasoning capabilities. This prompt us to further curate data using the same method in the future that shall bring forth even more competent models.

\begin{table*}[t]
\centering
\small
\begin{tabular}{l|lllllll}
\toprule
{Type}    & {LLaVA-1.5} & {+ \newshortname{}} & {(50\%)} & {(25\%)} & {(10\%)} & {(5\%)} & {(2.5\%)} \\
\midrule
add\_att         & 80.64              & \textbf{95.09}        & 94.22                & 90.90                & 88.58                & 88.44               & 86.27                 \\
add\_obj         & 87.83              & \textbf{97.82}        & 97.58                & 95.44                & 94.67                & 93.79               & 90.45                 \\
replace\_att     & 92.64              & \textbf{93.53}        & 93.27                & 91.50                & 92.51                & 91.62               & 92.64                 \\
replace\_obj     & 96.73              & \textbf{98.37}        & \textbf{98.37}       & 98.24                & 98.06                & 97.88               & 97.28                 \\
replace\_rel     & 87.13              & 85.92                 & 84.92                & 85.63                & 86.77                & 88.12               & \textbf{88.19}        \\
swap\_att        & 86.34              & \textbf{92.79}        & 90.54                & 89.19                & 88.44                & 90.09               & 89.79                 \\
swap\_obj        & 84.90              & \textbf{85.72}        & 83.67                & 81.63                & 82.04                & 84.08               & \textbf{86.12}        \\
add\_average     & 86.02              & \textbf{97.13}        & 96.73                & 94.29                & 93.13                & 92.44               & 89.39                 \\
replace\_average & 92.38              & 92.83                 & 92.40                & 92.24                & 92.79                & 93.02               & 93.00                 \\
swap\_average    & 85.95              & \textbf{90.89}        & 88.69                & 87.15                & 86.71                & 88.47               & 88.80                 \\
Overall          & 89.27              & \textbf{94.17}        & 93.54                & 92.38                & 92.18                & 92.26               & 91.17\\
\bottomrule
\end{tabular}
\caption{Ratio of augmentation ablation study on fine-tuning LLaVA-1.5.}
\label{tab:ratio-llava}
\end{table*}

\begin{table*}[t]
\centering
\small
\begin{tabular}{l|lllllll}
\toprule
{Type}    & {CLIP} & {+ \newshortname{}} & {(50\%)} & {(25\%)} & {(10\%)} & {(5\%)} & {(2.5\%)} \\
\midrule
add\_att         & 87.14         & \textbf{90.34}        & 89.23                & 88.11                & 86.85                & 87.09               & 87.05                 \\
add\_obj         & 77.89         & \textbf{86.70}        & 81.21                & 79.91                & 78.46                & 78.03               & 77.74                 \\
replace\_att     & 82.61         & \textbf{87.94}        & 84.77                & 82.99                & 83.37                & 82.86               & 82.61                 \\
replace\_obj     & 93.76         & \textbf{95.94}        & 94.79                & 94.37                & 93.88                & 93.76               & 93.76                 \\
replace\_rel     & 68.91         & \textbf{76.24}        & 72.97                & 70.19                & 69.55                & 69.34               & 69.13                 \\
swap\_att        & 67.41         & \textbf{73.57}        & 68.76                & 66.81                & 66.66                & 66.51               & 67.26                 \\
swap\_obj        & 60.00         & \textbf{68.57}        & 62.85                & 62.85                & 63.26                & 62.44               & 60.40                 \\
add\_average     & 80.21         & \textbf{87.62}        & 83.22                & 81.97                & 80.57                & 80.31               & 80.08                 \\
replace\_average & 82.39         & \textbf{87.10}        & 84.76                & 83.20                & 82.83                & 82.60               & 82.47                 \\
swap\_average    & 65.42         & \textbf{72.22}        & 67.17                & 65.75                & 65.75                & 65.42               & 65.42                 \\
Overall          & 79.54         & \textbf{85.49}        & \textbf{85.65}       & 82.07                & 80.64                & 79.94               & 79.68      \\
\bottomrule
\end{tabular}
\caption{Ratio of augmentation ablation study on fine-tuning CLIP.}
\label{tab:ratio-clip}
\end{table*}

\section{More evaluation on CLIP downstream tasks}
\label{sec:down-clip}
In Table~\ref{tab:down-clip}, we compare vanilla CLIP with \newshortname{}-finetuned CLIP on Flickr30k-Attributes on some common downstream tasks. The results show that \newshortname{} generally improves performance.
\begin{table*}[t]
\centering
\small
\begin{tabular}{l|ll}
\toprule
{Benchmark}                     & \newshortname{} & LAION-2b \\
\midrule
ImageNet-a \cite{Hendrycks2021-mf}                   & \textbf{26.8}          & 26.25                        \\
kitti-closest-vehicle-distance \cite{Ginger_undated-uz} & \textbf{30.1}          & 26.16                        \\
eurosat \cite{Helber2019-gi}                     & \textbf{51.59}         & 48.13                        \\
voc2007\_multilabel \cite{Everingham2015-lg}          & \textbf{82.49}         & 79.63                        \\
dmlab \cite{Zhai2019-aa}                      & \textbf{16.81}         & 15.97                        \\
pcam  \cite{Veeling2018-yr}                   & \textbf{60.58}         & 59.78                        \\
fgvc\_aircraft \cite{Maji2013-ju}           & \textbf{25.26}         & 24.54       \\
\bottomrule
\end{tabular}
\caption{Comparing our fine-tuned CLIP with LAION-2b pretrained CLIP on common downstream tasks.}
\label{tab:down-clip}
\end{table*}

\end{document}